\documentclass[10pt,twocolumn,letterpaper]{article}

\usepackage{wacv}              

\usepackage{graphicx}
\usepackage{amsmath}
\usepackage{amssymb}
\usepackage{booktabs}

\usepackage{multirow} 
\usepackage{pgfplots} 
\usepackage{colortbl} 
\usepackage{indentfirst} 
\usepackage{amssymb} 

\usepackage[pagebackref,breaklinks,colorlinks]{hyperref}

\usepackage[capitalize]{cleveref}
\crefname{section}{Sec.}{Secs.}
\Crefname{section}{Section}{Sections}
\Crefname{table}{Table}{Tables}
\crefname{table}{Tab.}{Tabs.}

\def\wacvPaperID{1755} 
\def\confName{WACV}
\def\confYear{2025}

\begin{document}

\title{\textbf{RT-DETRv3: Real-time End-to-End Object Detection with Hierarchical Dense Positive Supervision}}


\author
{ 
Shuo Wang\thanks{Equal Contribution.}
\quad
Chunlong Xia$^*$
\quad
Feng Lv
\quad
Yifeng Shi\thanks{Corresponding Author.}
\\
Baidu Inc, China
\\
\tt\small\{wangshuo36, xiachunlong, lvfeng02, shiyifeng\}@baidu.com
}

\maketitle
\thispagestyle{empty}

\begin{abstract}
RT-DETR is the first real-time end-to-end transformer-based object detector. Its efficiency comes from the framework design and the Hungarian matching. However, compared to dense supervision detectors like the YOLO series, the Hungarian matching provides much sparser supervision, leading to insufficient model training and difficult to achieve optimal results. To address these issues, we proposed a hierarchical dense positive supervision method based on RT-DETR, named RT-DETRv3. Firstly, we introduce a CNN-based auxiliary branch that provides dense supervision that collaborates with the original decoder to enhance the encoder's feature representation. Secondly, to address insufficient decoder training, we propose a novel learning strategy involving self-attention perturbation. This strategy diversifies label assignment for positive samples across multiple query groups, thereby enriching positive supervisions. Additionally, we introduce a shared-weight decoder branch for dense positive supervision to ensure more high-quality queries matching each ground truth. Notably, all aforementioned modules are training-only. We conduct extensive experiments to demonstrate the effectiveness of our approach on COCO val2017. RT-DETRv3 significantly outperforms existing real-time detectors, including the RT-DETR series and the YOLO series. For example, RT-DETRv3-R18 achieves 48.1\% AP (+1.6\%/+1.4\%) compared to RT-DETR-R18/RT-DETRv2-R18, while maintaining the same latency. Furthermore, RT-DETRv3-R101 can attain an impressive 54.6\% AP outperforming YOLOv10-X. The code will be released at \url{https://github.com/clxia12/RT-DETRv3}.



\end{abstract}

\begin{figure}[t]
\centering
\includegraphics[scale=0.85]{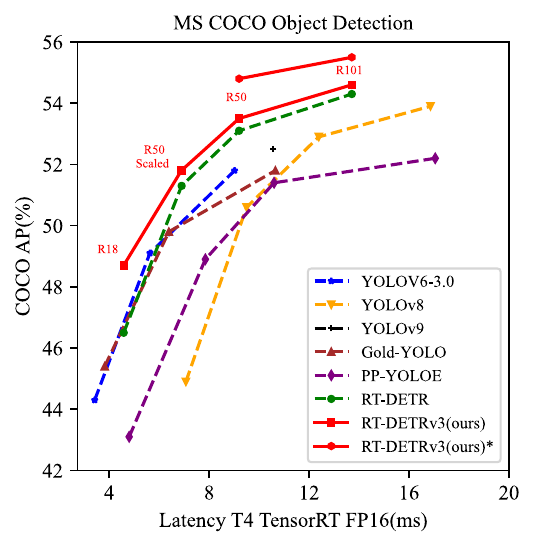}
\caption{\textbf{Compared to other real-time object detectors.} Our method has better performance in the trade-off between speed and accuracy. $*$ represents adding extra data.}
\label{latency}
\end{figure}

\section{Introduction}
\label{sec:intro}

Object detection is an important fundamental problem in computer vision, which mainly focuses on obtaining the position and category information of objects in the image. Real-time object detection has higher requirements for algorithm performance, such as inference speed greater than 30 FPS. It has enormous value in practical applications such as autonomous driving, video surveillance, and object tracking. In recent years, real-time object detections have garnered significant attention from both researchers and industry professionals due to its efficient inference speed and superior detection accuracy. Among these, the most popular are single-stage real-time object detectors based on CNNs, such as the YOLO series (~\cite{bochkovskiy2020yolov4, jocher2022yolov5, li2022yolov6, wang2023yolov7, yolov8, wang2024yolov9, wang2024yolov10}). They all adopted a one-to-many label assignment strategy, designed an efficient inference framework, and used non-maximum suppression (NMS) to filter redundant prediction results. Although this strategy introduced additional latency, they achieved a trade-off between accuracy and speed.

DETR~\cite{carion2020detr} is the first transformer-based end-to-end object detection algorithm. It employs set prediction and is optimized through the Hungarian matching strategy, eliminating the need for NMS post-processing and thereby simplifying the object detection process. Subsequent DETRs (such as DAB-DETR~\cite{liu2022Dab-detr}, DINO~\cite{zhang2022dino}, and DN-DETR~\cite{li2022Dn-detr}, etc.) further introduce iterative refinement schemes and denoising training,
which effectively accelerating the convergence speed of the model and improving its performance. 
However, its high computational complexity significantly limits its practical applications.

RT-DETR~\cite{zhao2024rt-detr} is the first real-time end-to-end transformer-based object detection algorithm. It designed an efficient hybrid encoder and IoU-aware query selection module, and a scalable decoder layer, achieving better results than other real-time detectors. However, the Hungarian matching strategy provides sparse supervision during training, leading to insufficient training of both the encoder and decoder, which limits the optimal performance of the approach. RT-DETRv2~\cite{lv2024rt-detrv2} further enhances the flexibility and practicality of RT-DETR~\cite{zhao2024rt-detr} by optimizing the training strategy to improve performance without sacrificing speed, although requires longer training time. 
To effectively address the issue of sparse supervision in object detection, 
we propose a hierarchical dense positive supervision method, which effectively accelerates model convergence and enhances model performance by introducing multiple auxiliary branches during training. Our main contributions are as follows:
\begin{itemize}
    \item We introduce a one-to-many label assignment auxiliary head based on CNN, which collaborates with the original detection branch for optimization, further enhancing the representational capability of the encoder.
    \item We propose a learning strategy with self-attention perturbations aimed at enhancing the supervision of the decoder by diversifying label assignments across multiple query groups. 
    Additionally, we introduced a shared-weight decoder branch for dense positive supervision to ensure more high-quality queries matching each ground truth. These approaches significantly improve the model's performance and accelerate convergence without additional inference latency.
    \item Extensive experiments conducted on the COCO dataset have thoroughly validated the effectiveness of our proposed approach. As shown in Figure~\ref{latency}, RT-DETRv3 significantly outperforms other real-time detectors, including the RT-DETR series and YOLO series. For instance, RT-DETRv3-R18 achieves 48.1\% AP (+1.6\%) compared to RT-DETR-R18, while maintaining the same latency. Additionally, RT-DETRv3-R50 outperforms YOLOv9-C by 0.9\% AP, even with a latency reduction of 1.3ms.
    
\end{itemize}
\begin{figure*}[t]
  \centering
  \includegraphics[scale=0.74]{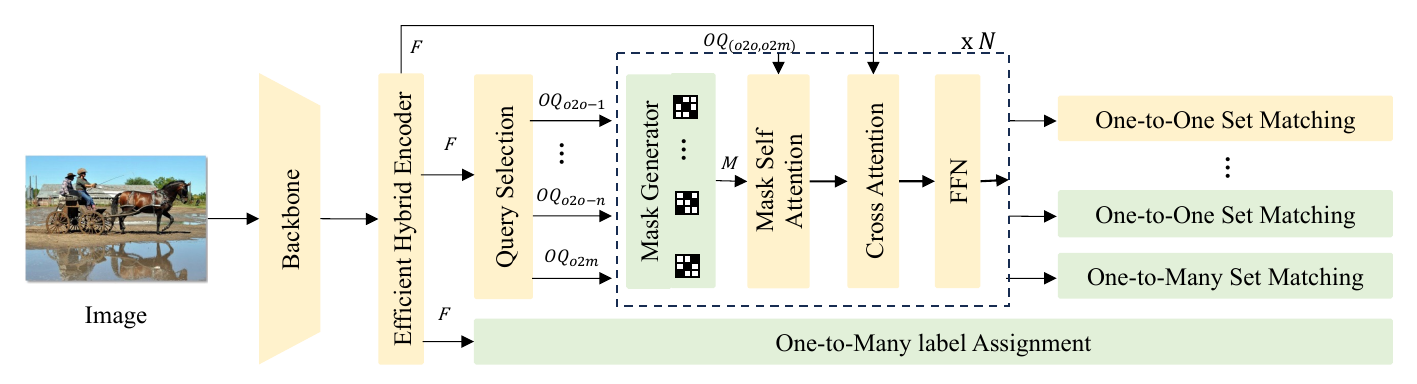}
  \caption{\textbf{Architecture of RT-DETRv3.} We preserve the core architecture of RT-DETR(highlighted in yellow)and propose a novel hierarchical decoupled dense supervision method (emphasized in green). Firstly, we enhance the encoder's representation capability by incorporating a CNN-based one-to-many label assignment auxiliary branch. Secondly, to enhance and strengthen supervision of the decoder, we generate multiple object queries (OQ) through the query selection module and apply random masking to perturb the self-attention mechanism, effectively diversifying the distribution of positive query samples. Additionally, to ensure that multiple relevant queries focus on the same target, we introduce a supplementary one-to-many matching branch. Notably, these auxiliary branches are discarded during evaluation.}
   \label{figure1}
\end{figure*}

\section{Related Work}
\subsection{CNN-based real-time object detection.}
The current CNN-based real-time object detectors are mainly the YOLO series. YOLOv4~\cite{bochkovskiy2020yolov4} and YOLOv5~\cite{jocher2022yolov5} optimized the network architecture (e.g., by adopting CSPNet~\cite{wang2020cspnet} and PAN~\cite{liu2018pan}), while also utilizing Mosaic data augmentation.
YOLOv6~\cite{li2022yolov6} further optimized the structure, including the RepVGG~\cite{ding2021repvgg} backbone, decoupled head, SimSPPF, and more effective training strategy (e.g., SimOTA~\cite{ge2021yolox-simOTA}, etc.). YOLOv7~\cite{wang2023yolov7} introduces the E-ELAN attention module to better integrate features from different levels and adopts the adaptive anchor mechanism to improve small object detection. YOLOv8~\cite{yolov8} proposed a C2f module for effective feature extraction and fusion. YOLOv9~\cite{wang2024yolov9} proposed a new GELAN architecture and designed a PGI to enhance the training process. The PP-YOLO series~\cite{long2020ppyolo, huang2021ppyolov2} is a real-time object detection solution based on the PaddlePaddle framework proposed by Baidu. This series of algorithms has been optimized and improved on the basis of the YOLO series, aiming to improve detection accuracy and speed to meet the needs of practical application scenarios.

\subsection{Transformer-based real-time object detection.}
RT-DETR~\cite{zhao2024rt-detr} is the first real-time end-to-end object detector. This approach designs an efficient hybrid encoder that effectively processes multi-scale features by decoupling intra-scale interactions and cross-scale fusion and proposes IoU-aware query selection to further improve performance by providing higher-quality initial object queries to the decoder. Its accuracy and speed are superior to the YOLO series of the same period, and it has received widespread attention. RT-DETRv2~\cite{lv2024rt-detrv2} further optimized the training strategy, including dynamic data augmentation and optimized sampling operators for easy deployment, resulting in further improvement of its model performance. However, due to their one-to-one sparse supervision, the convergence speed and final effect are limited. Therefore, introducing a one-to-many label assignment strategy can further improve the model's performance.

\subsection{Auxiliary training strategy.}
Co-DETR~\cite{zong2023co-detr} proposed multiple parallel one-to-many label assignment auxiliary head training strategies (e.g., ATSS~\cite{zhang2020ATSS} and Faster RCNN~\cite{ren2015faster-rcnn}), which can easily enhance the learning ability of the encoder in end-to-end detectors. For example, the integration of ViT-CoMer~\cite{xia2024vit-comer} with Co-DETR~\cite{zong2023co-detr} has achieved state-of-the-art performance on the COCO detection task. DAC-DETR~\cite{hu2024DAC-DETR}, MS-DETR~\cite{zhao2024ms-detr}, and GroupDETR~\cite{chen2023group-detr} mainly accelerate the convergence of the model by adding one-to-many supervised information to the decoder of the model. The above approaches accelerate the convergence or improve the performance of the model by adding additional auxiliary branches at different positions of the model, but they are not real-time object detectors. Inspired by these, we introduced multiple one-to-many auxiliary dense supervision modules to both the encoder and decoder of RT-DETR~\cite{zhao2024rt-detr}. These modules enhance the convergence speed and improve the overall performance of the RT-DETR~\cite{zhao2024rt-detr}. Since these modules are only involved during the training phase, they don't affect the inference latency of RT-DETR~\cite{zhao2024rt-detr}.

\section{Method}
\subsection{Overall Architecture.}
The overall structure of RT-DETRv3 is shown in Figure~\ref{figure1}. We have retained the overall framework of RT-DETR~\cite{zhao2024rt-detr} (highlighted in yellow) and additionally introduced our proposed hierarchical decoupling dense supervision method (highlighted in green). Initially, the input image is processed through a CNN backbone (e.g., ResNet~\cite{he2016resnet}) and a feature fusion module, termed the efficient hybrid encoder, to obtain multi-scale features \{$C_{3}$, $C_{4}$, and $C_{5}$\}. These features are then fed into a CNN-based one-to-many auxiliary branch and a transformer-based decoder branch in parallel. For the CNN-based one-to-many auxiliary branch, we directly employ existing state-of-the-art dense supervision methods, such as PP-YOLOE~\cite{xu2022ppyoloe}, to collaboratively supervise the encoder's representation learning. In the transformer-based decoder branch, the multi-scale features are first flattened and concatenated. We then use a query selection module to select the top-k features from them to generate object queries. Within the decoder, we introduce a mask generator that produces multiple sets of random masks. These masks are applied to the self-attention module, affecting the correlation between queries and thus differentiating the assignments of positive queries. Each set of random masks is paired with a corresponding query, as depicted in the Figure~\ref{figure1} by $OQ_{o2o-1},..., OQ_{o2o-n}$. Furthermore, to ensure that there are more high-quality queries matching each ground truth, we incorporate an one-to-many label assignment branch within the decoder. The following sections provide a detailed description of the modules proposed in this work.

\subsection{Overview of RT-DETR.}
RT-DETR~\cite{zhao2024rt-detr} is a real-time detection framework designed for object detection tasks. It integrates the advantages of end-to-end prediction from DETR~\cite{2020detr} while optimizing inference speed and detection accuracy. To achieve real-time performance, the encoder module is replaced with a lightweight CNN backbone, and an Efficient Hybrid Encoder module which designed for efficient feature fusion. RT-DETR~\cite{zhao2024rt-detr} proposed an Uncertainty-minimal query selection module to select high-confidence feature as object queries, reducing the difficulty of query optimization. Subsequently, multiple layers of the decoder enhance these queries through self-attention, cross-attention and feed-forward network (FFN) modules, with the prediction results produced by MLP layers. During the training optimization process, RT-DETR~\cite{zhao2024rt-detr} employs Hungarian matching for one-to-one assignment. For loss calculation, it uses L1 loss and GIoU loss to supervise box regression, and variable focus loss (VFL) to supervise the learning of the classification task.

\subsection{One-to-Many Auxiliary Branch Based on CNN.}
To alleviate the problem of sparse supervision in encoder output caused by the decoder's one-to-one set matching scheme, we introduce an auxiliary detection head with one-to-many assignment, such as PP-YOLOE~\cite{xu2022ppyoloe}. This strategy can effectively strengthen the supervision of the encoder, enabling it to have sufficient representation ability to accelerate the convergence of the model. Specifically, we directly integrate the output features \{$C_{3}$, $C_{4}$, and $C_{5}$\} of the encoder into PP-YOLOE head, For the one-to-many matching algorithm, we follow the configuration of the PP-YOLOE head and use the ATSS matching algorithm in the early stage of training, and then switch to the TaskAlign matching algorithm. For the learning of classification and localization tasks, VFL and distributed focus loss (DFL) were respectively selected. Among them, VFL uses IoU scores as the target for positive samples, which makes positive samples with high IoU contribute relatively more to the loss. This also makes the model focus more on high-quality samples rather than low-quality samples during training. Specifically, decoder head also use VFL loss to ensure consistency in task definition. We denote the overall loss of the CNN auxiliary branch as $L_{aux}$, with the corresponding loss weight denoted as $\alpha$.

\begin{figure}[t]
  \centering
  \includegraphics[scale=0.85]{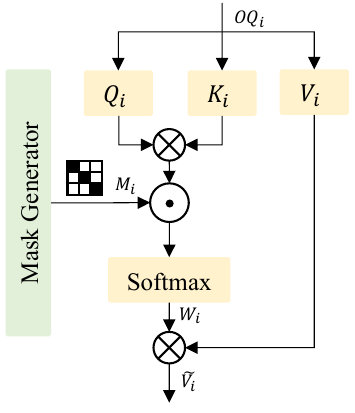}
   \caption{\textbf{Mask self-attention module.} $M_{i}$ represents the perturbation mask corresponding to the $i$-$th$ set of object queries. $\otimes$ denotes matrix multiplication, and $\odot$ denotes element-wise multiplication.}
   \label{figure3}
\end{figure}

\subsection{Multi-Group Self-Attention Perturbation Branches Based on Transformer.}

The decoder consists of a series of transformer blocks, with each block incorporating a self-attention, cross-attention, and FFN (Feed-Forward Network) module. Initially, the queries interact with each other through the self-attention module to enhance or diminish their feature representations. Subsequently, each query updates itself by retrieving information from the encoder's output features via the cross-attention module. Lastly, the FFN predicts the class and bounding box coordinates of the target corresponding to each query. However, the adoption of a one-to-one set matching in the RT-DETR leads to sparse supervision information, ultimately impairing the model's performance.

To ensure that multiple related queries associated with the same target have the opportunity to participate in positive sample learning, we propose multiple self-attention perturbation modules based on Mask Self-Attention. The implementation details of this perturbation module are shown in Figure~\ref{figure1}. First, we generate multiple sets of object queries through the query selection module, denoted as $OQ_{i}$ ($i$=$1...N$, where $N$ is the number of sets). Correspondingly, we use a mask generator to generate a random perturbation mask $M_{i}$ for each set of $OQ_{i}$. Both $OQ{i}$ and $M_{i}$ are fed into the Mask Self-Attention module, resulting in the perturbed and fused features.

The detailed implementation of the Mask Self-Attention module is shown in Figure~\ref{figure3}, $OQ_{i}$ is first linearly projected to obtain $Q_{i}$, $K_{i}$, and $V_{i}$. Then, $Q_{i}$ and $K_{i}$ are multiplied to compute the attention weight, which is further multiplied by $M_{i}$ and passed through a softmax function to yield the perturbed attention weight. Finally, this perturbed attention weight is multiplied by ${V}_{i}$ to obtain the fused result $\tilde{V}_{i}$ . The process can be represented as:
\begin{equation}
    Q_{i}, K_{i}, V_{i} = Linear(OQ_{i})
\end{equation}
\begin{equation}
    W_{i} = Softmax(M_{i}(Q_{i}K^{T}_{i}))
\end{equation}
\begin{equation}
    \tilde{V}_{i} = W_{i}V_{i}
\end{equation}
The introduction of multiple sets of random perturbations diversifies the features of the queries, allowing multiple related queries associated with the same target to have a chance of being assigned as positive sample queries, thereby enriching the supervision information. During training, multiple sets of object queries are concatenated and fed into a single decoder branch, enabling parameter sharing and enhancing training efficiency. The loss computation and label assignment scheme remain consistent with RT-DETR. We denote the loss of the $i$-$th$ set as $Loss^{i}_{o2o}$, and the total loss for N perturbation sets is calculated as follow:
\begin{equation}
L_{o2o} = \frac{1}{N} \sum_{1}^{N}L^{i}_{o2o} 
\end{equation}
with the corresponding loss weight denoted as $\beta$.

\begin{table*}[]
\centering
\setlength{\tabcolsep}{7.5pt} 
\renewcommand{\arraystretch}{1.3}
\begin{tabular}{l|c|c|c|c|ccc}
\hline
\multicolumn{1}{c|}{Model}               & Backbone & \#Params(M) & GFLOPs & Latency (ms)  & $AP^{val}$ 1x & $AP^{val}$ 3x & $AP^{val}$ 6x \\ \hline
RT-DETR~\cite{zhao2024rt-detr}           & \multirow{3}{*}{R18}         & \multirow{3}{*}{20}            & \multirow{3}{*}{60}      & \multirow{3}{*}{4.6}          & 38.7          & 44.5          & 46.5          \\
RT-DETRv2~\cite{lv2024rt-detrv2}         &      &           &      &       & 39.8          & 44.9          & 46.7/47.9\textsuperscript{$\dagger$}          \\
\rowcolor[HTML]{ECF4FF} RT-DETRv3~(ours) &          &             &        &           & \textbf{41.5} & \textbf{46.1} & \textbf{48.1/48.7\textsuperscript{$\dagger$}} \\ \hline
RT-DETR~\cite{zhao2024rt-detr}          & \multirow{3}{*}{R34}         & \multirow{3}{*}{31}            & \multirow{3}{*}{92}      & \multirow{3}{*}{6.3}           & 42.8          & 47.5          & 48.9          \\
RT-DETRv2~\cite{lv2024rt-detrv2}         &       &           &      &       & 43.0          & 47.2          & 49.0/49.9\textsuperscript{$\dagger$}          \\
\rowcolor[HTML]{ECF4FF} RT-DETRv3~(ours) &          &             &        &           & \textbf{44.7} & \textbf{48.6} & \textbf{49.9/50.1\textsuperscript{$\dagger$}} \\ \hline
RT-DETR~\cite{zhao2024rt-detr}           & \multirow{3}{*}{R50}         & \multirow{3}{*}{42}            & \multirow{3}{*}{136}      & \multirow{3}{*}{9.2}           & 48.9          & 52.2          & 53.1          \\
RT-DETRv2~\cite{lv2024rt-detrv2}         &       &           &     &        & -             & -             & 53.4          \\
\rowcolor[HTML]{ECF4FF} RT-DETRv3~(ours) &          &             &        &           & \textbf{50.2} & \textbf{53.0}   & \textbf{53.4} \\ \hline
RT-DETR~\cite{zhao2024rt-detr}          & \multirow{3}{*}{R101}         & \multirow{3}{*}{76}            & \multirow{3}{*}{259}      & \multirow{3}{*}{13.5}           & 50.2          & 53.5          & 54.3          \\
RT-DETRv2~\cite{lv2024rt-detrv2}         &        &          &     &        & -             & -             & 54.3          \\
\rowcolor[HTML]{ECF4FF} RT-DETRv3~(ours) &          &             &        &           & \textbf{51.3} & \textbf{54.2} & \textbf{54.6} \\ \hline
\end{tabular}
\caption{\textbf{Comparison of the RT-DETR series for object detection on COCO val2017.} R18, R34, R50, and R101 refer to ResNet-18, ResNet-34, ResNet-50, and ResNet-101, respectively. 1x, 3x, and 6x respectively correspond to training for 12, 36, and 72 epochs. $\dagger$ denotes training for 120 epochs.}

\label{speed_transformer}
\end{table*}

\subsection{One-to-Many Dense Supervision Branch Based on Transformer.}

To maximize benefits of multi-group self-attention perturbation branches, we introduce an additional dense supervision branch with shared weights in the decoder. This ensures more high-quality queries matching each ground truth. Specifically, we employ a query selection module to generate a unique set of object queries.
During the sample matching phase, an augmented target set is generated by replicating the training labels by a factor of $m$, with a default value of 4. This augmented set is subsequently matched against the prediction of the query. The loss computation remains consistent with the original detection loss, and we designate $L_{o2m}$ as the loss function for this branch, with a loss weight of $\gamma$.

\subsection{Total Loss.}
In summary, the overall loss function of our proposed approach is as follows: 
\begin{equation}
    L = \alpha L_{aux} + \beta L_{o2o} + \gamma L_{o2m}
\end{equation}
where $L_{aux}$ is responsible for dense supervision of the encoder, $L_{o2o}$ enriches the one-to-one supervision information for the decoder while preserving the end-to-end prediction characteristics, and $L_{o2m}$ provides one-to-many dense supervision to the decoder. By default, the loss weights $\alpha$, $\beta$, and $\gamma$ are set to 1.

\section{Experiments}

\subsection{Datasets and Evaluation Metrics.}
We selected the MS COCO 2017~\cite{lin2014coco} object detection dataset as the evaluation benchmark for our approach. This dataset consists of 115k training images and 5k test images. We adopted the same evaluation metric, AP, as used in the RT-DETR~\cite{zhao2024rt-detr} approach. 
We compared the performance of RT-DETRv3 with other real-time object detectors in terms of convergence efficiency, inference speed, and effectiveness, which include both transformer-based and CNN-based real-time object detectors.
Additionally, we conducted ablation studies on the modules mentioned in this paper. All experimental details and results are elaborated in the following sections.

\begin{table*}[t]
\centering
\setlength{\tabcolsep}{7.5pt} 
\renewcommand{\arraystretch}{1.25}
\begin{tabular}{l|cccc|c}

\hline
\multicolumn{1}{c|}{Model}                    & \#Epochs & \#Params (M) & GFLOPs & Latency (ms) & $AP^{val}$            \\ \hline
YOLOv6-3.0-S~\cite{li2022yolov6}              & 300       & 18.5         & 45.3   & 3.4          & 44.3          \\
Gold-YOLO-S~\cite{wang2024gold-yolo}          & 300       & 21.5         & 46.0   & 3.8          & 45.4          \\
YOLO-MS-S~\cite{chen2023yolo-ms}              & 300       & 8.1          & 31.2   & 10.1         & 46.2          \\
YOLOv8-S~\cite{yolov8}                        & 500       & 11.2         & 28.6   & 7.1          & 46.2          \\
YOLOv9-S~\cite{wang2024yolov9}                & 500       & 7.1          & 26.4   & -            & 46.7          \\
YOLOV10-S~\cite{wang2024yolov10}              & 500       & 7.2          & 21.6   & 2.5          & 46.3          \\
\rowcolor[HTML]{ECF4FF} RT-DETRv3-R18~(ours)  & 120       & 20           & 60     & 4.6          & \textbf{48.7} \\ \hline
YOLOv6-3.0-M~\cite{li2022yolov6}              & 300       & 34.9         & 85.8   & 5.6          & 49.1          \\
Gold-YOLO-M~\cite{wang2024gold-yolo}          & 300       & 41.3         & 87.5   & 6.4          & 49.8          \\
YOLO-MS~\cite{chen2023yolo-ms}                & 300       & 22.2         & 80.2   & 12.4         & 51.0          \\
YOLOv8-M~\cite{yolov8}                        & 500       & 25.9         & 78.9   & 9.5          & 50.6          \\
YOLOv9-M~\cite{wang2024yolov9}                & 500       & 20.0         & 76.3   & -            & 51.1          \\
YOLOV10-M~\cite{wang2024yolov10}              & 500       & 15.4         & 59.1   & 4.7          & 51.1          \\
\rowcolor[HTML]{ECF4FF} RT-DETRv3-R34~(ours)  & 120       & 31           & 92     & 6.3          & 50.1          \\
\rowcolor[HTML]{ECF4FF} RT-DETRv3-R50m~(ours) & 72        & 36           & 100    & 6.89         & \textbf{51.7} \\ \hline
Gold-YOLO-L~\cite{wang2024gold-yolo}          & 300       & 75.1         & 151.7  & 9.0          & 51.8          \\
YOLOv5-X~\cite{jocher2022yolov5}              & 300       & 86           & 205    & 23.3         & 50.7          \\
PPYOLOE-L~\cite{xu2022ppyoloe}                & 300       & 52           & 110    & 10.6         & 51.4          \\
YOLOv6-L~\cite{li2022yolov6}                  & 300       & 59           & 150    & 10.1         & 52.8          \\
YOLOv7-L                                      & 300       & 36           & 104    & 18.2         & 51.2          \\
YOLOV8-L~\cite{yolov8}                        & 500       & 43           & 165    & 14.1         & 52.9          \\
YOLOv9-C~\cite{wang2024yolov9}                & 500       & 25.3         & 102.1  & 10.57        & 52.5          \\
YOLOV10-L~\cite{wang2024yolov10}              & 500       & 24.4         & 120.3  & 7.28         & 53.2          \\
\rowcolor[HTML]{ECF4FF} RT-DETRv3-R50~(ours)  & 72        & 42           & 136    & 9.2          & \textbf{53.4} \\ \hline
YOLOv8-X~\cite{yolov8}                        & 500       & 68.2         & 257.8  & 16.9         & 53.9          \\
YOLOv10-X~\cite{wang2024yolov10}              & 500       & 29.5         & 160.4  & 10.7         & 54.4          \\
\rowcolor[HTML]{ECF4FF} RT-DETRv3-R101~(ours) & 72        & 76           & 259    & 13.5         & \textbf{54.6} \\ \hline
\end{tabular}
\caption{\textbf{Compared to CNN-based real-time object detectors on COCO val2017.}}
\label{performance based on cnn}
\end{table*}

\subsection{Implementation Details.}

We integrated proposed hierarchical dense supervision branches into the RT-DETR~\cite{zhao2024rt-detr} framework. The CNN-based dense supervision auxiliary branch directly employed the PP-YOLOE head, with its sample matching strategy, loss calculation, and all other configurations consistent with those of PP-YOLOE~\cite{xu2022ppyoloe}. We reused the RT-DETR~\cite{zhao2024rt-detr} decoder structure as the main branch and additionally added three groups of parameter-shared self-attention perturbation branches. The sample matching method is consistent with the main branch, utilizing Hungarian matching algorithm. We also added a parameter-shared one-to-many matching branch, where each ground truth is matched with four object queries by default and set 300 object queries in total. The AdamW optimizer, integrated with a weight decay factor of 0.0001, was employed, ensuring that all other training configurations adhered strictly to the RT-DETR\cite{zhao2024rt-detr}, encompassing both data augmentation and pre-training. We used a 10x (120 epochs) training schedule for smaller backbones (R18, R34) and a 6x (72 epochs) training schedule for larger backbones (R50, R101). We use four NVIDIA A100 GPUs to train our proposed method with a batch size of 64. Moreover, the latencies of all models are tested on T4 GPU with TensorRT FP16, following ~\cite{zhao2024rt-detr}. We have observed that, in comparison to most detectors employing longer training epochs, RT-DETRv3 only needs 72 epochs to achieve superior accuracy.

\subsection{Comparison with Transformer-based Real-time Object Detectors.}
\textbf{Inference speed and algorithm performance.} The real-time object detectors based on transformer architecture are primarily represented by the RT-DETR series. Table~\ref{speed_transformer} presents the comparison results between our approach and the RT-DETR series. Our approach outperforms both RT-DETR~\cite{zhao2024rt-detr} and RT-DETRv2~\cite{lv2024rt-detrv2} across various backbone. Specifically, compared to RT-DETR~\cite{zhao2024rt-detr}, with the 6x training schedule, our approach demonstrates improvements of 1.6\%, 1.0\%, 0.3\%, and 0.3\% with the R18, R34, R50, and R101 backbones, respectively. In comparison to RT-DETRv2~\cite{lv2024rt-detrv2}, we evaluated the R18 and R34 backbones under 6x/10x training schedules, where our approach improvements of 1.4\%/0.8\% and 0.9\%/0.2\%, respectively. Moreover, since the auxiliary dense supervision branches we proposed are training-only, our approach maintains the same inference speed as both RT-DETR~\cite{zhao2024rt-detr} and RT-DETRv2~\cite{lv2024rt-detrv2}. 

\begin{table}[]
\centering
\setlength{\tabcolsep}{5.25pt} 
\renewcommand{\arraystretch}{1.2}
\begin{tabular}{c|c|c|c}
\hline
Method                           & Extra data & Epochs & $AP^{val}$ \\ \hline
\multirow{4}{*}{RT-DETRv3-R50}   & x          & 51     & 53.4       \\
                                 & x          & 72     & 52.9~\textcolor{red}{(-0.5)}       \\
  & \checkmark & 51     & 54.2       \\
                                 & \checkmark & 72     & 54.8~\textcolor{blue}{(+0.6)}        \\ \hline
\multirow{4}{*}{RT-DETRv3-R101}  & x          & 51     & 54.6       \\
                                 & x          & 72     & 54.2~\textcolor{red}{(-0.4)}       \\ 
 & \checkmark & 51     & 54.7       \\
                                 & \checkmark & 72     & 55.4~\textcolor{blue}{(+0.7)}       \\ \hline
\end{tabular}


\caption{\textbf{Analysis of overfitting.} Extra data represents the Object365 dataset~\cite{shao2019obj365}.}
\label{Extra_data}
\end{table}

\textbf{Convergence speed.}~Our approach builds on the RT-DETR~\cite{zhao2024rt-detr} framework by incorporating CNN-based and transformer-based one-to-many dense supervision, which not only boosts model performance but also speeds up convergence. We have conducted extensive experiments to validate the effectiveness of our approach. Table ~\ref{speed_transformer} presents a comparative analysis of RT-DETRv3, RT-DETR~\cite{zhao2024rt-detr}, and RT-DETRv2~\cite{lv2024rt-detrv2} across various training schedules. It clearly demonstrates that our method outperforms them in terms of convergence speed in any schedule and only needs half of the training epochs to achieve the comparable performance.

\begin{figure}[t]
\centering
  \includegraphics[scale=0.75]{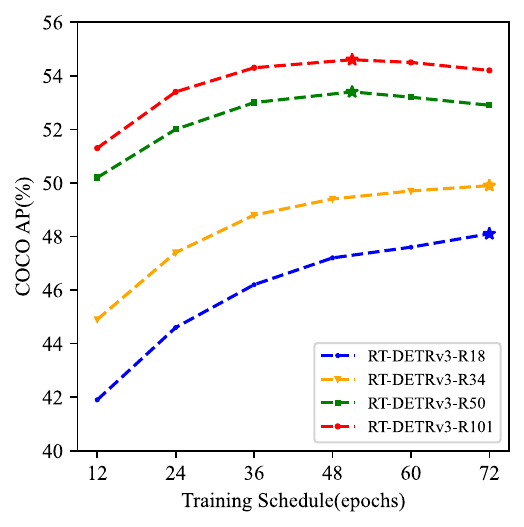}
\caption{\textbf{Convergence curves of RT-DETRv3 across different model sizes. } $\star$ represents the best AP.}
\label{figure4}
\end{figure}

\textbf{Analysis of overfitting.} As illustrated in Figure~\ref{figure4}, we noticed that as the model size increases, RT-DETRv3 tends to exhibit overfitting. We believe this may be due to a mismatch between the size of the training dataset and the model size. We conducted several experiments, as shown in Table~\ref{Extra_data}, when we added additional training data, the performance of RT-DETRv3 continues to improve as the training epochs increase, and it performs better than the model without the additional data at the same epochs.


\subsection{Comparison with CNN-Based Real-time Object Detectors.}
\textbf{Inference speed and algorithm performance.} ~We compared the end-to-end speed and accuracy of RT-DETRv3 with current advanced CNN-based real-time object detection methods. We categorized the models into small, medium, and large scales based on their inference speed. Under similar inference performance conditions, we compared RT-DETRv3 with other state-of-the-art algorithms such as YOLOv6-3.0~\cite{li2022yolov6}, Gold-YOLO~\cite{wang2024gold-yolo}, YOLO-MS~\cite{chen2023yolo-ms}, YOLOv8~\cite{yolov8}, YOLOv9~\cite{wang2024yolov9}, and YOLOv10~\cite{wang2024yolov10}. As shown in Table~\ref{performance based on cnn}, for small-scale models, the RT-DETRv3-R18 approach outperforms YOLOv6-3.0-S, Gold-YOLO-S, YOLO-MS-S, YOLOv8-S, YOLOv9-S, and YOLOv10-S by 4.4\%, 3.3\%, 2.5\%, 2.5\%, 2.0\%, and 2.4\%, respectively. For medium-scale models, RT-DETRv3 also demonstrates superior performance compared to YOLOv6-3.0-M, Gold-YOLO-M, YOLO-MS-M, YOLOv8-M, YOLOv9-M, and YOLOv10-M. For large-scale models, our approach consistently outperforms CNN-based real-time object detectors. For example, our RT-DETRv3-R101 can achieve 54.6 AP, which is better than YOLOv10-X. However, since we have not yet optimized the overall framework of the RT-DETRv3 detector for lightweight deployment, there is still room for further improving the inference efficiency of RT-DETRv3.

\textbf{Convergence speed.} ~As shown in Table~\ref{performance based on cnn}, we are excited to find that our RT-DETRv3, while achieving superior performance, can reduce the training epochs to as little as 60\% or even less compared to CNN-based real-time detectors.

\subsection{Ablation Study.}

\textbf{Settings.}~We conducted the ablation experiments using RT-DETR~\cite{zhao2024rt-detr} as the baseline and then validated the impact of our proposed approach by sequentially integrating auxiliary CNN-based one-to-many label assignments branch, the auxiliary transformer-based one-to-many label assignments branch, and the multi-group self-attention perturbation modules. These experiments were performed with ResNet18 as the backbone, with a batch size of 64, and four NVIDIA A100 GPUs, while maintaining other configurations consistent with RT-DETR~\cite{zhao2024rt-detr}.

\textbf{Ablation for components.}~We conducted ablation experiments to evaluate proposed modules in this paper. As shown in Table~\ref{Ablation}, each module significantly enhances the model's performance. For instance, by adding O2M-T module to RT-DETR~\cite{zhao2024rt-detr}, we observed a 1.0\% improvement in performance over the base model. When all proposed modules are integrated into RT-DETR for algorithm optimization, the model's performance improves by 1.6\%.

\begin{table}[]
\centering
\renewcommand{\arraystretch}{1.2}
\begin{tabular}{l|ccc|c}
\hline
\multicolumn{1}{c|}{Method} & O2M-C      & O2M-T      & MGSA       & $AP^{val}$            \\ \hline
                            & x          & x          & x          & 46.5          \\
                            & \checkmark & x          & x          & 47.4          \\
RT-DETR                     & x          & \checkmark & x          & 47.5          \\
                            & x          & x          & \checkmark & 47.5          \\
\rowcolor[HTML]{ECF4FF}     & \checkmark & \checkmark & \checkmark & \textbf{48.1} \\ \hline
\end{tabular}
\caption{\textbf{Ablation studies of key components.} O2M-C represents the one-to-many auxiliary branch based on CNN, O2M-T refers to one-to-many dense supervision branch based on transformer, and MGSA stands for multi-group self-attention perturbation branch based on transformer.}
\label{Ablation}
\end{table}

\textbf{Number of self-attention perturbation branches.} To verify the effect varying the number of self-attention perturbation branches on RT-DETRv3 performance, we conducted ablation experiments using RT-DETRv3-R18 with branch counts of 2, 3, and 4, while keeping all other configurations unchanged. As shown in Table~\ref{Ablation_number}, when the number of branches was set to 3, the model achieved its optimal performance with AP 48.1. Reducing the number of branches decreased the richness of the supervision signals, leading to lower performance. Conversely, increasing the number of branches excessively raised the model's learning difficulty without yielding significant performance gains.

\begin{table}[]
\centering
\renewcommand{\arraystretch}{1.2}
\begin{tabular}{c|cc}
\hline
Number & $AP^{val}$ & $AP^{val}_{50}$ \\ \hline
2      & 47.9 & 65.4      \\
3      & 48.1 & 65.6      \\
4      & 48.0 & 65.3      \\ \hline
\end{tabular}
\caption{\textbf{Ablation study on the number of self-attention perturbation branches.}}
\label{Ablation_number}
\end{table}

\section{Conclusion}

In this paper, we propose a real-time object detection algorithm based on transformer, named RT-DETRv3. This algorithm builds upon RT-DETR by incorporating multiple dense positive sample auxiliary supervision modules. These modules apply one-to-many object supervision to specific features of both the encoder and decoder in RT-DETR, thereby accelerating the algorithm's convergence and improving its performance. It's important to note that these modules are training-only. We validated the effectiveness of our algorithm on the COCO object detection benchmark, and the experiments demonstrate that our algorithm achieves better results compared to other real-time object detectors. We hope that our work can inspire researchers and developers working on real-time transformer-based object detection.





{\small
\bibliographystyle{ieee_fullname}
\bibliography{main.bbl}
}

\end{document}